\documentclass[10pt,twocolumn,letterpaper]{article}

\usepackage{cvpr}
\usepackage{times}
\usepackage{epsfig}
\usepackage{graphicx}
\usepackage{amsmath}
\usepackage{amssymb}
\usepackage{enumerate}
\usepackage{enumitem}

\usepackage{url}            
\usepackage{booktabs}       
\usepackage{amsfonts}       
\usepackage{nicefrac}       
\usepackage{microtype}      
\usepackage{caption}
\usepackage{subcaption}
\usepackage{booktabs,dcolumn}

\usepackage[pagebackref=true,breaklinks=true,letterpaper=true,colorlinks,bookmarks=false]{hyperref}

\cvprfinalcopy 

\def\cvprPaperID{1939} 

\ifcvprfinal\fi
\begin{document}

\title{Boundary Flow: A Siamese Network that Predicts Boundary Motion without Training on Motion}

\author{Peng Lei, Fuxin Li and Sinisa Todorovic\\
Oregon State University\\
Corvallis, OR 97331, USA\\
{\tt\small \{leip, lif, sinisa\}@oregonstate.edu}
}

\maketitle

\begin{abstract}
Using deep learning, this paper addresses the problem of joint object boundary detection and boundary motion estimation in videos, which we named boundary flow estimation. Boundary flow is an important mid-level visual cue as boundaries characterize objects' spatial extents, and the flow indicates objects' motions and interactions. Yet, most prior work on motion estimation has focused on dense object motion or feature points that may not necessarily reside on boundaries. For boundary flow estimation, we specify a new fully convolutional Siamese network (FCSN) that jointly estimates object-level boundaries in two consecutive frames. Boundary correspondences in the two frames are predicted by the same FCSN with a new, unconventional deconvolution approach. Finally, the boundary flow estimate is improved with an edgelet-based filtering. Evaluation is conducted on three tasks: boundary detection in videos, boundary flow estimation, and optical flow estimation. On boundary detection, we achieve the state-of-the-art performance on the benchmark VSB100 dataset. On boundary flow estimation, we present the first results on the Sintel training dataset. For optical flow estimation, we run the recent approach CPM-Flow but on the augmented input with our boundary-flow matches, and achieve significant performance improvement on the Sintel benchmark.
\end{abstract}

\begin{figure}[ht]
\begin{center}
\includegraphics[width=1.0\linewidth]{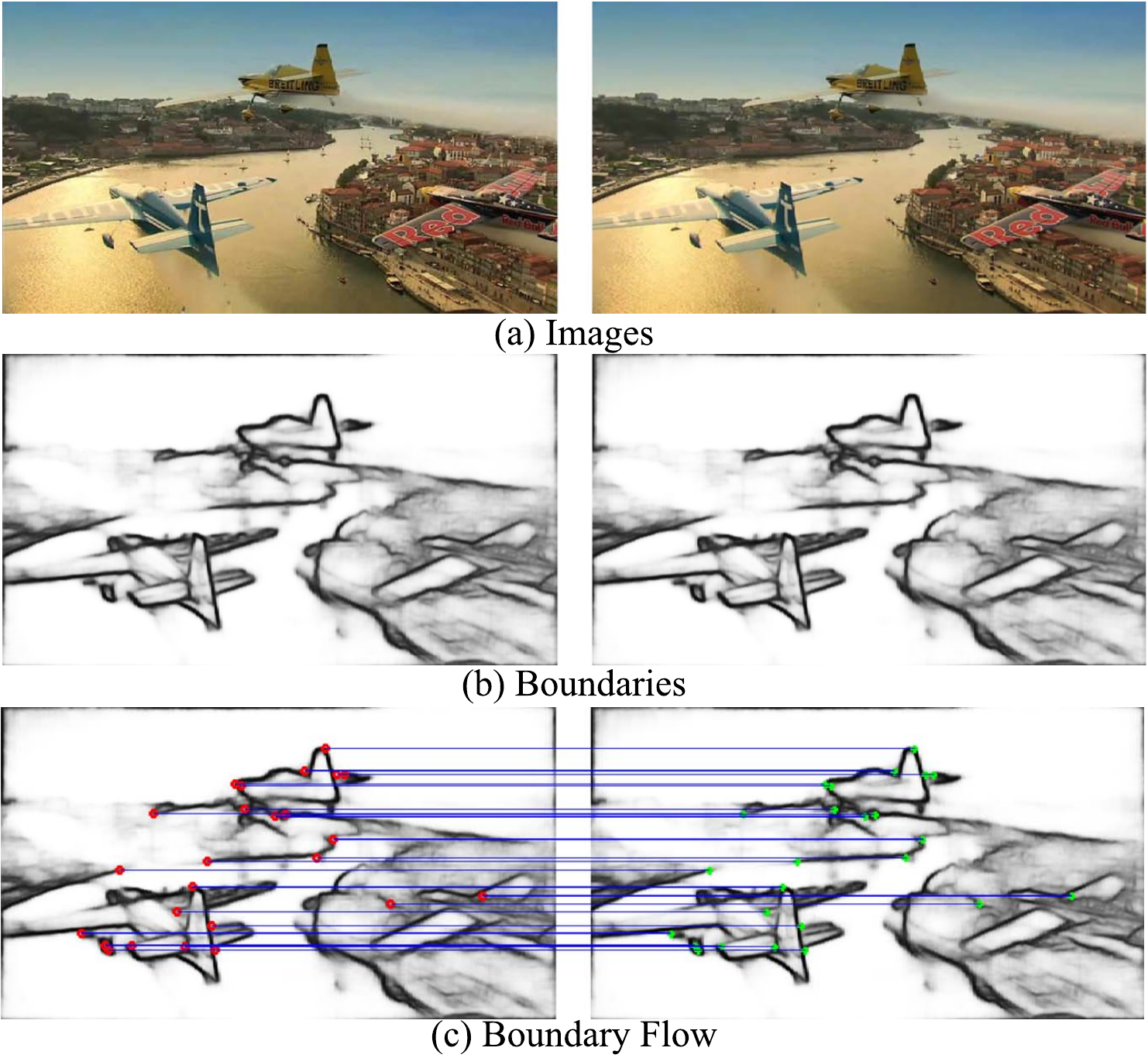}
\end{center}
\vskip -0.2in
   \caption{Boundary flow estimation. Given two images (a), our approach {\em jointly}: predicts object boundaries in both images (b), and estimates motion of the boundaries in the two images (c). For clarity, only a part of boundary matches are shown in (c).}
\vskip -0.05in
\label{fig:illustration}
\end{figure}

\section{Introduction}

This paper considers the problem of estimating motions of object boundaries in two consecutive video frames, or simply two images. We call this problem boundary flow (BF) estimation. Intuitively, BF is defined as the motion of every pixel along object boundaries in two images, as illustrated in Fig.~\ref{fig:illustration}. A more rigorous definition will be presented in Sec.~\ref{sec:BoundaryFlow}.
BF estimation is an important problem. Its solution can be used as an informative mid-level visual cue for a wide range of higher-level vision tasks, including object detection~(e.g.,\cite{ferrari2008groups}), object proposals~(e.g.,\cite{zitnick2014edge}), video segmentation~(e.g.,\cite{lee2011key}), and depth prediction~(e.g.,~\cite{amer2015monocular}). This is because, in a BF, the boundaries identify objects' locations, shapes,  motions, local interactions, and figure-ground relationships. In many object-level tasks, BF can be computed in lieu of the regular optical flow, hence avoiding estimating motion on many irrelevant background pixels that may not be essential to the performance of the task.

Yet, this problem has received scant attention in the literature. 
Related work has mostly focused on single-frame edge detection and dense optical flow estimation. These approaches, however, cannot be readily applied to BF estimation, due to new challenges. In particular, low-level spatiotemporal boundary matching --- which is agnostic of objects, scenes, and motions depicted in the two video frames --- is subject to many ambiguities. 
The key challenge is that distinct surfaces sharing a boundary move with different motions,  out-of-plane rotations and changing occlusions. This makes appearance along the boundary potentially inconsistent in consecutive frames.
The difficulty of matching
boundaries in two images also increases when multiple
points along the boundary have similar appearance.

Our key hypothesis is that because of the rich visual cues along the boundaries, BF may be learned \textbf{without} pixel-level motion annotations, 
which is typically very hard to come by (prior work resorts to simulations~\cite{mayer2016large} or computer graphics~\cite{ButlerECCV2012}, which may not represent realistic images).

\begin{figure*}
\begin{center}
\includegraphics[width=.75\linewidth]{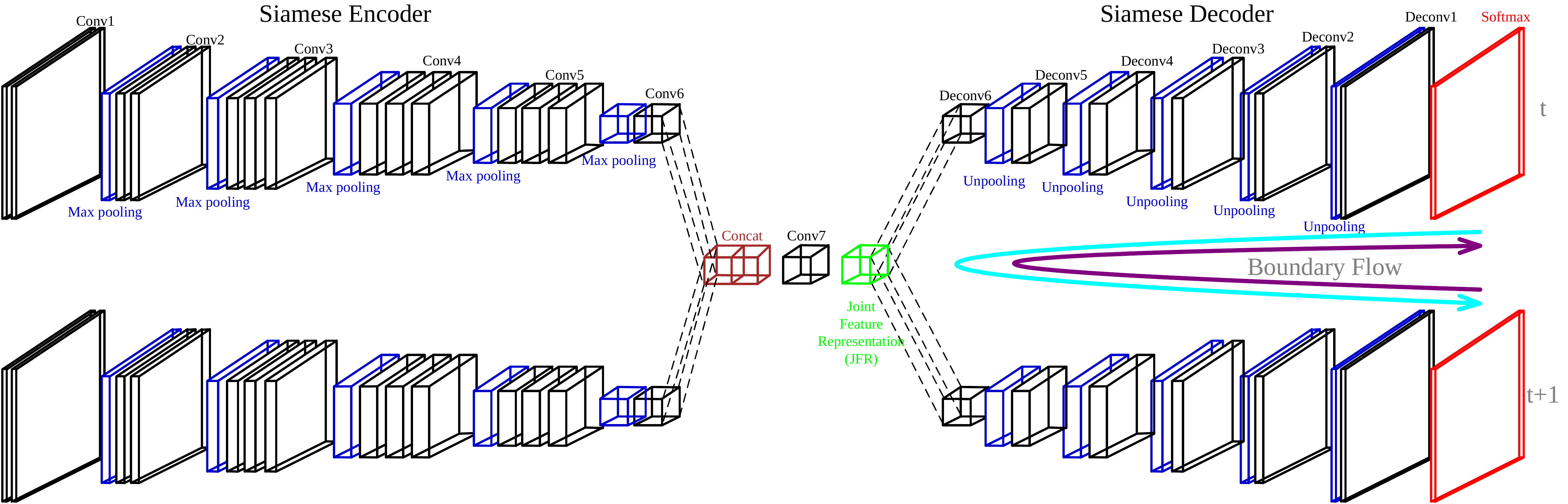}
\end{center}
\vskip -0.15in
   \caption{FCSN consists of a Siamese encoder and a Siamese decoder and takes two images as input. The two Siamese soft-max outputs of the decoder produce boundary predictions in each of the two input images. 
   Also, the decoder associates the two Siamese branches via the decoder layers and the JFR layer (the green cube) for 
   calculating the excitation attention score, which in turn is used for BF estimation, as indicated by the cyan and purple arrows. 
   The convolution, pooling, softmax and concatenation layers are marked with black, blue, red and brown respectively. Best viewed in color. 
   }
   \vskip -0.15in
\label{fig:arch}
\end{figure*}

While there are a few approaches that separately detect and match boundaries in a video, e.g., \cite{liu2006analysis, thompson1998exploiting, thompson1985dynamic}, to the best of our knowledge, this is the first work that gives a rigorous definition of boundary flow, as well as {\em jointly} detects object boundaries and estimates their flow within the deep learning framework.
We extend ideas from deep boundary detection approaches in images~\cite{xie2015holistically,yang2016object}, and specify a new Fully Convolutional Siamese encoder-decoder Network (FCSN) for joint spatiotemporal boundary detection and BF estimation. As shown in Fig.~\ref{fig:arch}, FCSN encodes two consecutive video frames into a coarse joint feature representation (JFR) (marked as a green cube in Fig.~\ref{fig:arch}). Then, a Siamese decoder uses deconvolution and un-max-pooling to estimate boundaries in each of the two input images. 

Our network trains only on boundary annotations in one frame and predicts boundaries in each frame, so at first glance it does not provide motion estimation. However, the Siamese network is capable of predicting different (but correct) boundaries in two frames, while the only difference in the two decoder branches are max-pooling indices. Thus, our key intuition is that there must be a common edge representation in the JFR layer for each edge, that are mapped to two different boundary predictions by different sets of max-pooling indices. 
Such a common representation enables us to match the corresponding boundaries in the two images. The matching is done  by 
tracking a boundary from one boundary prediction image back to the JFR, and then from the JFR to boundaries in the other boundary prediction image. 
This is formalized as an excitation attention-map estimation of the FCSN. 
We use edgelet-based matching to further improve the smoothness and enforce ordering of pixel-level boundary matching along an edgelet.

Since FCSN performs boundary detection and provides correspondence scores for boundary matching, we say that FCSN {\em unifies} both boundary detection and BF estimation within the same deep architecture. In our experiments, this approach proves capable of handling large object displacements in the two images, and thus can be used as an important {\em complementary} input to dense optical flow estimation.

We evaluate FCSN on the VSB100  dataset \cite{NB13} for boundary detection, and on the 
Sintel training dataset \cite{ButlerECCV2012} for BF estimation. Our results demonstrate that 
FCSN yields higher precision on boundary detection than the state of the art, and 
using the excitation attention score for boundary matching yields superior BF performance relative 
to reasonable baselines. Also, experiments performed on the Sintel test dataset show that we can use 
the BF results to augment the input of a state-of-the-art optical flow algorithm -- CPM-Flow~\cite{huefficient16} --
and generate significantly better dense optical flow than the original.

Our key contributions are summarized below:
\begin{itemize}[itemsep=-5pt,topsep=2pt, partopsep=1pt]
\item We consider the problem of BF estimation within the deep learning framework, give a rigorous definition of BF,  and specify and extensively evaluate a new deep architecture FCSN for solving this problem. We also demonstrate the utility of BF for estimating dense optical flow.
\item We propose a new approach to generate excitation-based correspondence scores from FCSN for boundary matching, and develop an edgelet-based matching for refining point matches along corresponding boundaries.
\item We improve the state-of-the-art on spatiotemporal boundary detection, provide the first results on BF estimation, and achieve competitive improvements on dense optical flow when integrated with CPM-Flow \cite{huefficient16}.
\end{itemize}

\section{Related Work}
This section reviews closely related work on boundary detection and dense optical flow estimation. The
literature on semantic video segmentation and semantic contour detection is beyond our scope.

\noindent\textbf{Boundary Detection}. Traditional approaches to boundary detection typically extract a multitude of hand-designed features at different scales, 
and pass them to a detector for boundary detection \cite{arbelaez2011contour}.  Some of these methods leverage the structure of local contour maps for fast edge detection \cite{Dollar2015PAMI}. Recent work resort to convolutional neural
networks (CNN) for learning deep features that are suitable for boundary detection \cite{ganin2014n, 
shen2015deepcontour, bertasius2015high, bertasius2015deepedge, xie2015holistically, 
yang2016object, maninis2016convolutional}. \cite{khoreva2016improved} trains a boundary detector on a video dataset and achieved improved results. Their network is defined on a single frame and does not provide motion information
across two frames. 
The approach of \cite{yang2016object} is closest to ours, since they use a fully convolutional encoder-decoder for boundary detection on one frame. However, without a Siamese network their work cannot be used to estimate boundary motion as proposed in this paper.

\noindent\textbf{Optical flow estimation.} There has been considerable efforts to improve the efficiency and robustness of optical flow estimation, including 
PatchMatch \cite{barnes2009patchmatch} and extensions \cite{korman2011coherency, he2012computing, bailer2015flow}. They compute the Nearest Neighbor Field (NNF) 
by random search and propagation. EpicFlow \cite{revaud2015epicflow} uses DeepMatching \cite{weinzaepfel2013deepflow} for a hierarchical matching of image patches, 
and its extension Coarse-to-fine Patch-Match (CPM-Flow) \cite{huefficient16} introduces a propagation between levels of the hierarchical matching. 
While EpicFlow~\cite{revaud2015epicflow} propagates optical flow to image boundaries, it still does not handle very abrupt motions well, 
as can be seen in many of the fast-moving objects in the Sintel benchmark dataset. In this paper, we do not focus on dense optical flow estimation, 
but demonstrate the capability of boundary flow estimation in supplementing optical flow, which is beneficial in large displacements and flow near boundaries. 
As our results show, we improve CPM-Flow when using our boundary flow estimation as a pre-processing step. Boundary motion estimation was first considered in \cite{liu2006analysis}, and then in \cite{xue2015computational} where dense optical flow was initialized from an optical flow computed on Canny edges. However, in both of these papers, the definition of their edge flow differs from our boundary flow in the following. First, they do not consider cases when optical flow is not defined. Second, they do not have a deep network to perform boundary detection. Finally, they do not evaluate edge flow as a separate problem.

\section{Boundary Flow}\label{sec:BoundaryFlow}
This section defines BF, 
introduces the FCSN, and specifies finding boundary correspondences in the two frames using the FCSN's excitation attention score.

\subsection{Definition of Boundary Flow}\label{definition_of_bf}
BF is defined as the motion of every boundary pixel towards the corresponding boundary pixel in the next frame. In the case of out-of-plane rotations and occlusions, BF identifies the occlusion boundary closest to the original boundary pixel (which becomes occluded).
We denote the set of boundaries in frame $t$ and $t+1$ as $B_1$ and $B_2$, respectively. Let $\text{OF}(\mathbf{x})$ denote the optical flow of a pixel $\mathbf{x}$ in frame $t$, and $\mathbf{x}+\text{OF}(\mathbf{x})$ represent a mapping of pixel $\mathbf{x}$ in frame $t+1$.  Boundary flow $\text{BF}(\mathbf{x})$ is defined as:

\noindent (\textit{i}) $\text{BF}(\mathbf{x}) = \arg \min_{\mathbf{y} \in B_2} \| \mathbf{y} - (\mathbf{x} + \text{OF}(\mathbf{x})) \|_2 - \mathbf{x}$, if $\text{OF}(\mathbf{x})$ exists;

\noindent (\textit{ii}) $\text{BF}(\mathbf{x}) = \text{OF}(\arg \min_{\mathbf{y}, \exists \text{OF}(\mathbf{y}) } \|\mathbf{y} - \mathbf{x}\|_2)$, if $\text{OF}(\mathbf{x})$ does not exist ($\mathbf{x}$ occluded in frame $t+1$);

\noindent (\textit{iii}) $\text{BF}(\mathbf{x})$ is undefined if \textit{argmin}  in (\textit{i}) or (\textit{ii})  does not return a unique solution.

\begin{figure}[t]
\begin{center}
   \includegraphics[width=.95\linewidth]{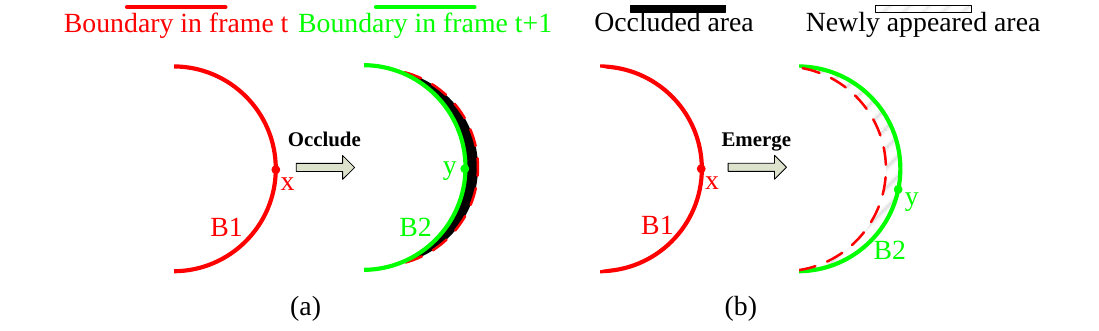}
\end{center}
\vskip -0.15in
   \caption{Fig.~\ref{fig:def_bf}(a) shows the case when a boundary $B_1$ in frame $t$ is occluded at time $t+1$.  Fig.~\ref{fig:def_bf}(b) shows the case when a boundary $B_1$ in frame $t$ is no longer a boundary at time $t+1$ but its pixels are visible. 
   In both cases BF is well-defined and always resides on the boundary.
}
\label{fig:def_bf}
\end{figure}

In (\textit{i}), BF is defined as optical flow for translations and elastic deformations, or the closest boundary pixel from the optical flow for out-of-plane rotations (see Fig.~\ref{fig:def_bf}(b)). In (\textit{ii}), BF is defined as the closest occlusion boundary of the pixel which becomes occluded (see Fig.~\ref{fig:def_bf}(a)). Thus, BF can be defined even if optical flow is not defined. Since optical flow is often undefined in the vicinity of occlusion boundaries, BF  captures shapes/occlusions better than optical flow. 
In (\textit{iii}), BF is undefined only in rare cases of fast movements with symmetric occluders (e.g. a perfect ball) resulting in multiple pixels as the argmin solution.

\subsection{Fully Convolutional Siamese Network}

We formulate boundary detection as a binary labeling problem. For this problem, we develop a new, end-to-end trainable FCSN, shown in Fig.~\ref{fig:arch}. FCSN takes two images as input, and produces binary soft-max outputs of boundary predictions in each of the two input images. The fully convolutional architecture in FCSN scales up to arbitrary image sizes.

FCSN consists of two modules: a Siamese encoder, and a Siamese decoder. The encoder stores all the pooling indices and encodes the two frames as the joint feature representation (JFR) (green box in Fig.~\ref{fig:arch}) through a series of convolution, ReLU, and pooling layers. The outputs of the encoder are concatenated, and then used as the input to the decoder. The decoder takes both the JFR and the max-pooling indices from the encoder as inputs. Then, the features from the decoder are passed into a softmax layer to get the boundary labels of all pixels in the two images.

The two branches of the encoder and the two branches of the decoder use the same architecture and share weights with each other. However, for two different input images, the two branches would still output different predictions, 
since decoder predictions are modulated with different pooling indices recorded in their corresponding encoder branches. Each encoder branch uses the layers of VGG net \cite{simonyan2014very} until the \emph{fc6} layer. The decoder decodes the JFR to the original input size through a set of unpooling, deconvolution, ReLU and dropout operations. Unlike the deconvolutional net \cite{noh2015learning} which uses a symmetric decoder as the encoder, we design a light-weight decoder with fewer weight parameters than a symmetric structure for efficiency. Except for the layer right before the \emph{softmax} layer, 
all the other convolution layers of the decoder are followed by a ReLU operator and a dropout layer.  
A detailed description of the convolution and dropout layers is summarized in Tab.~\ref{table:decoder}.

 \begin{table}[h]
 \begin{center}
 \begin{tabular}{|c|l|c|}
 \hline
 Layer & Filter & Dropout rate \\
 \hline\hline
 Deconv1 & $1 \times 1 \times 512$ & 0.5\\
 \hline
 Deconv2 & $5 \times 5 \times 512$ & 0.5\\
 \hline
 Deconv3 & $5 \times 5 \times 256$ & 0.5\\
 \hline
 Deconv4 & $5 \times 5 \times 128$ & 0.5\\
 \hline
 Deconv5 & $5 \times 5 \times 64$  & 0.5\\
 \hline
 Deconv6 & $5 \times 5 \times 32$  & 0.5\\
 \hline
 Softmax & $5 \times 5 \times 1$   & - \\
 \hline
 \end{tabular}
   \caption{The configuration of the decoder in FCSN.}
 \label{table:decoder}
 \end{center}
 \end{table}

\begin{figure*}[t]
\begin{center}
\includegraphics[width=1.0\linewidth]{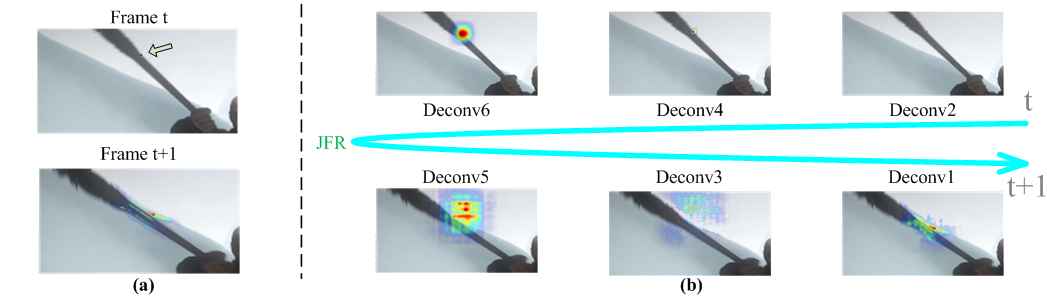}
\end{center}
\vskip -0.15in
   \caption{(a) Estimation of the excitation attention score in frame $t+1$ (bottom) for a particular boundary point in frame $t$ (top; the point is indicated by the arrow).  The attention map is well-aligned with the corresponding boundary in frame $t+1$, despite significant motion. 
   (b) Visualization of attention maps at different layers of the decoders of FCSN along the excitation path (cyan) 
   from a particular boundary point in frame $t$ to frame $t+1$ via the JFR. 
   For simplicity, we only show the attention maps in some of the 
layers from the decoder branch at time $t$ and $t+1$. As can be seen, starting from a pixel on the predicted boundary in frame $t$, the attention map gradually becomes coarser along the path to the JFR. Then from the JFR to boundary prediction in frame $t+1$, the excitation attention scores gradually become refined and more focused on the most relevant pixels in frame $t+1$. (Best viewed in color) 
}
\label{fig:genfeatures}
\end{figure*}

\subsection{Boundary Flow Estimation}
This section first describes estimation of the excitation attention score, used as a cue for boundary matching, and then 
specifies our edgelet-based matching for refining point matches along the boundaries.


\subsubsection{Excitation Attention Score}\label{sec:ExcitationBP}
A central problem in BF estimation is to identify the correspondence between a pair of boundary points $\langle\mathbf{x}_{t}^i,\mathbf{y}_{t+1}^j\rangle$, where $\mathbf{x}_{t}^i$ is a boundary point in frame $t$, and $\mathbf{y}_{t+1}^j$ is a boundary point in frame $t+1$. 
Our key idea is to estimate this correspondence by computing the excitation attention scores in frame $t+1$ for every  $\mathbf{x}_{t}^i$ in frame $t$, as well as the excitation attention scores in frame $t$ for every  $\mathbf{y}_{t+1}^j$ in frame $t+1$. The excitation attention scores can be generated efficiently using excitation backpropagation (ExcitationBP) \cite{zhang2016top} -- a probabilistic winner-take-all approach that models dependencies of neural activations through convolutional layers of a neural network for identifying relevant neurons for prediction, i.e., attention maps.  

The intuition behind our approach is that the JFR stores a joint representation of two corresponding boundaries of the two images, and thus could be used as a ``bridge'' for matching them. This ``bridge'' is established by tracking the most relevant neurons along the path from one branch of the decoder to the other branch via the JFR layer (the cyan and purple arrows in Fig.~\ref{fig:arch}).

In our approach, the winner neurons are sequentially sampled for each layer 
on the path from frame $t$ to $t+1$ via the JFR, based on a conditional winning probability. The relevance of each neuron is defined as its probability of being selected as a winner on the path. Following \cite{zhang2016top}, we define the  winning probability of a neuron
$a_m$ as 
\begin{align}
p(a_m)&=\sum_{n \in \mathcal{P}_m} p(a_m|a_n) p(a_n)&&\\\nonumber
      &=\sum_{n \in \mathcal{P}_m}\frac{w_{mn}^{+}a_m}{\sum_{m' \in \mathcal{C}_n}w_{m'n}^{+}a_{m'}}p(a_n)&&
\label{eq:MWP}
\end{align}

where 
$w_{mn}^{+} = max\{0, w_{mn}\}$, 
$\mathcal{P}_m$ and $\mathcal{C}_n$ denote the parent nodes of $a_m$ and the set of children of $a_n$ in the path traveling order, respectively. 
For our path  that goes from the prediction back to the JFR layer, $\mathcal{P}_m$ refers to all neurons in the layer closer to the prediction, and $\mathcal{C}_n$ refers to all neurons in the layer closer to the JFR layer.

ExcitationBP efficiently identifies which neurons are responsible for the final prediction.
In our approach, ExcitationBP can be run in parallel for each edgelet (see next subsection) of a predicted boundary. 
Starting from boundary predictions in frame $t$, we compute the marginal winning probability of all neurons along the path to the JFR. 
Once the JFR is reached, these probabilities are forward-propagated in the decoder branch of FCSN for 
finally estimating the pixel-wise excitation attention scores in frame $t+1$. For a pair of boundary points, we obtain the attention score $s_{i \rightarrow j}$.
Conversely, starting from boundary predictions in frame $t+1$, 
we compute the marginal winning probability of all neurons along the path to JFR, 
and feed them forward through the decoder for computing the excitation attention map 
in frame $t$. Then we can obtain the attention score $s_{j \rightarrow i}$. The attention score between a pair of boundary points 
$\langle\mathbf{x}_{t}^i,\mathbf{y}_{t+1}^j\rangle$ is defined as the average of $s_{i \rightarrow j}$ and $s_{j \rightarrow i}$, which we denote as $s_{ij}$.
An example of our ExcitationBP is shown in Fig.~\ref{fig:genfeatures}.

\subsubsection{Edgelet-based Matching} 
After estimating the excitation attention 
scores $s_{ij}$ of boundary point pairs $\langle\mathbf{x}_{t}^i,\mathbf{y}_{t+1}^j\rangle$, as described in Sec.~\ref{sec:ExcitationBP}, 
we use them for matching corresponding boundaries that have been predicted in frames $t$ and $t+1$. While there are many boundary matching methods that would be suitable, 
in this work we use the edgelet-based matching which not only finds good boundary correspondences, but also produces the detailed point matches along the boundaries, 
as needed for our BF estimation. To this end, we first decompose the predicted boundaries into smaller edgelets, then apply edgelet-based matching to pairs of edgelets.

\begin{figure}
\begin{center}
\includegraphics[width=1.\linewidth]{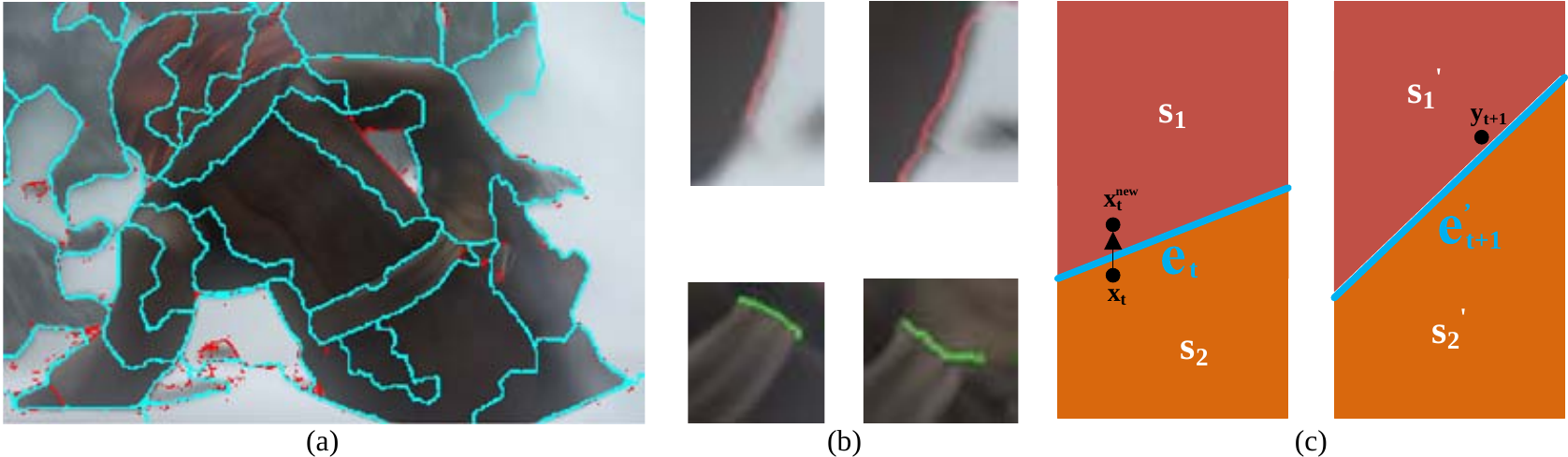}
\end{center}
\vskip -0.15in
   \caption{
Overview of edgelet matching. The matching process consists of three phases: superpixel generation, edgelet matching, and
flow placement. The two frames are first over-segmented into large superpixels using the FCSN boundaries. (a) most of the
boundary points (in red color) are well aligned with the superpixel boundaries (in cyan color); (b) Example edgelet matches. In the second case, it can be seen clearly that the appearance only
matches on one side of the edgelet. (c) The process of matching and flow placement. Sometimes, because of the volatility of edge detection, $\mathbf{x}_{t}$ and $\mathbf{y}_{t+1}$ falls on different sides of the boundary, we will need to then move $\mathbf{x}_{t}$ so that they fall on the same side. Note that $s_1$ and $s_2$, $s_1^{'}$ and $s_2^{'}$
denote the superpixel pairs falling on the two sides of the edgelets.  
}
\label{fig:matching}
\end{figure}

\noindent\textbf{From predicted boundaries to edgelets}. Given the two input images and their boundary predictions from FCSN, we oversegment the two frames using sticky superpixels \cite{Dollar2015PAMI}, 
and merge the superpixels to larger regions as in \cite{humayun2015middle}. Importantly, 
both oversegmentation and superpixel-merging use our boundary predictions as input, 
ensuring that contours of the resulting regions strictly respect our predicted boundaries, as illustrated in Fig.~\ref{fig:matching}(a).
We define an edgelet as all the points that lie on a given boundary shared by a 
pair of superpixels. 
Fig.~\ref{fig:matching}(b) shows two examples of matching edgelet pairs in frames $t$ and $t+1$.

\begin{figure*}
\begin{center}
\includegraphics[width=.85\linewidth]{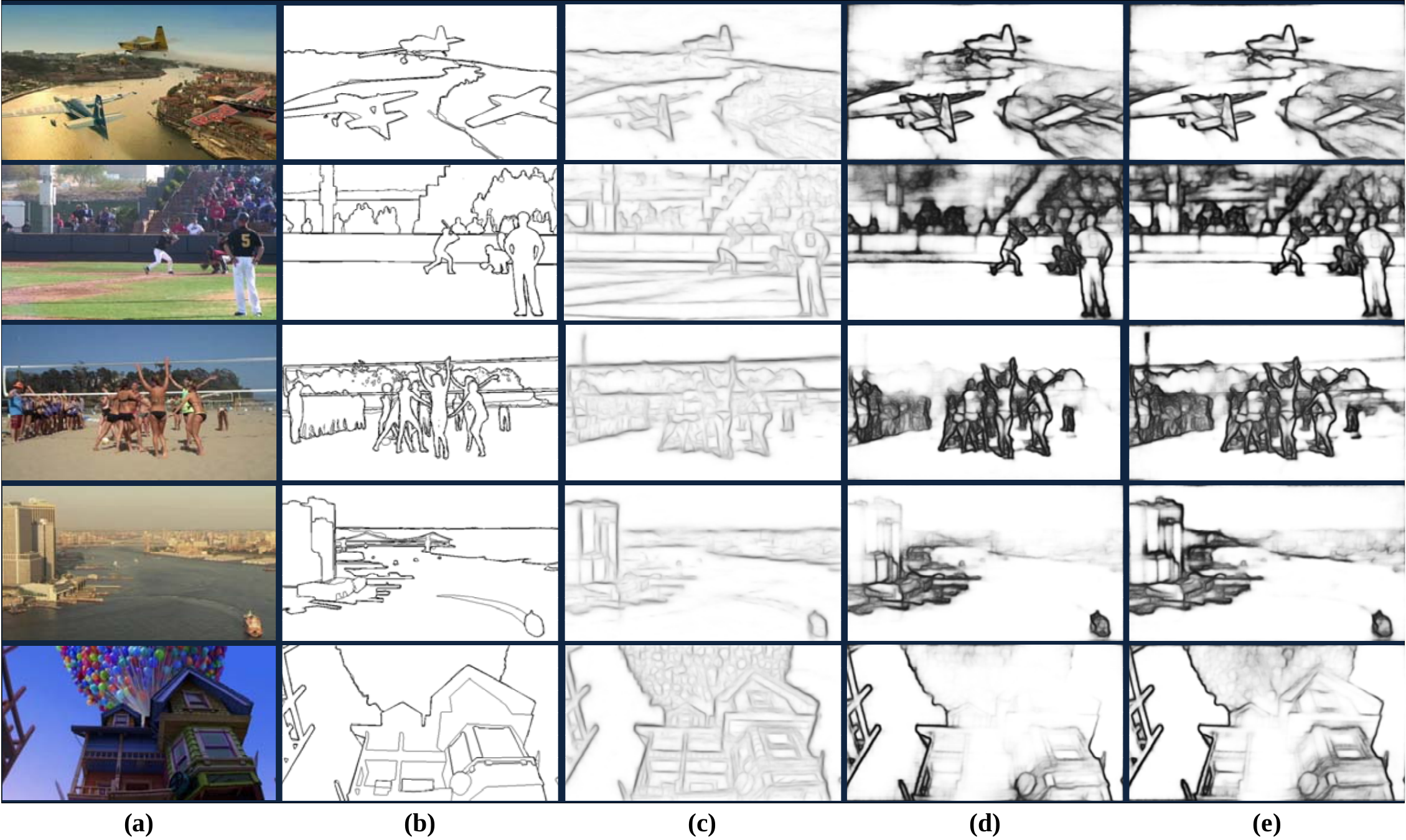}
\end{center}
\vskip -0.15in
   \caption{Example results on VSB100. In each row from left to right we present (a) input image, (b) ground truth annotation,
(c) edge detection~\cite{Dollar2015PAMI}, (d) object contour detection~\cite{yang2016object} and (e) our boundary detection.}
\label{fig:boundary}
\end{figure*}

\begin{figure}[htb]
\begin{center}
        \begin{subfigure}[b]{0.50\columnwidth}
                \includegraphics[width=\linewidth]{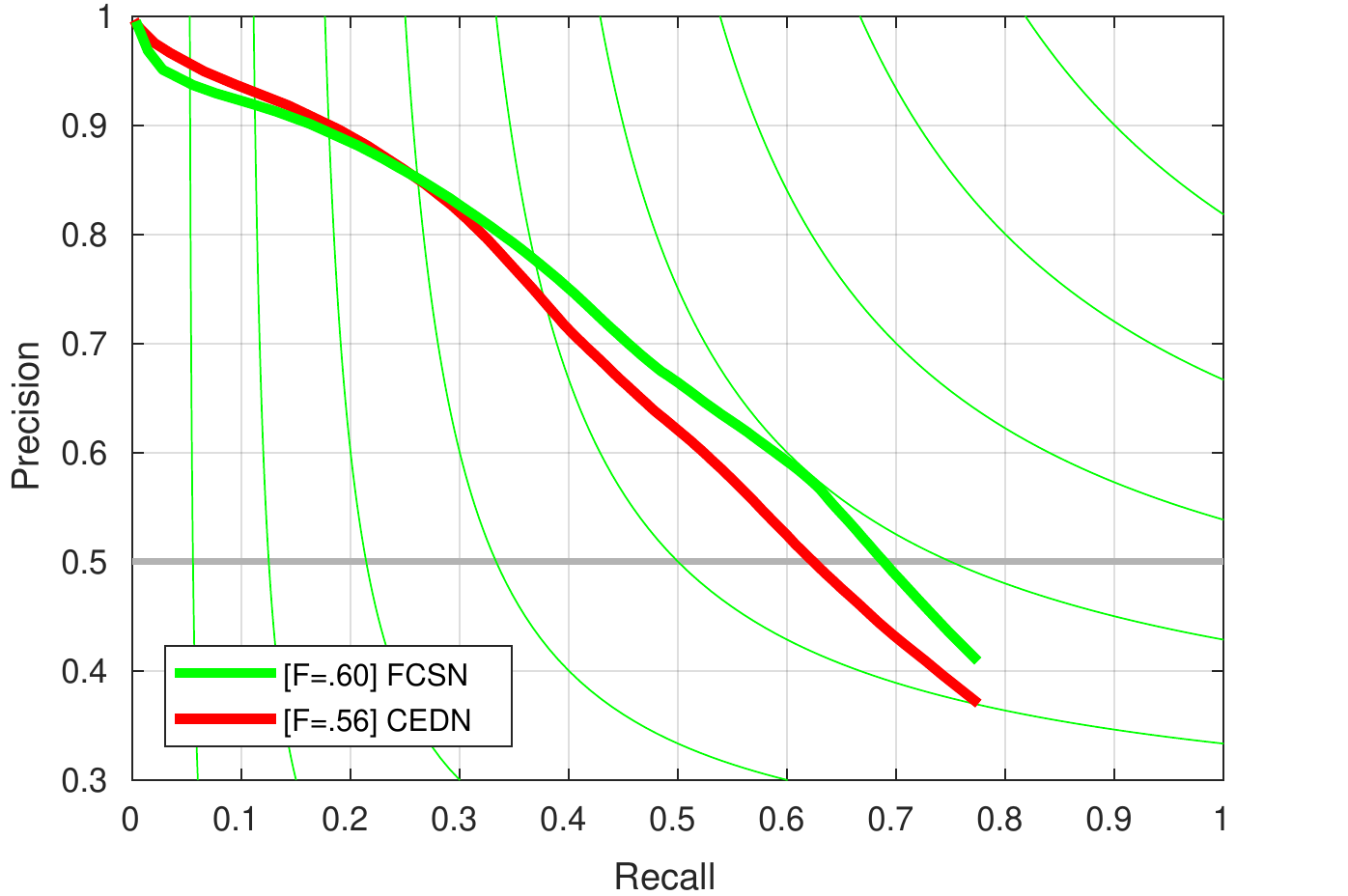}
                \caption{}
                \label{fig:cednvsfcsn}
        \end{subfigure}%
        \begin{subfigure}[b]{0.50\columnwidth}
                \includegraphics[width=\linewidth]{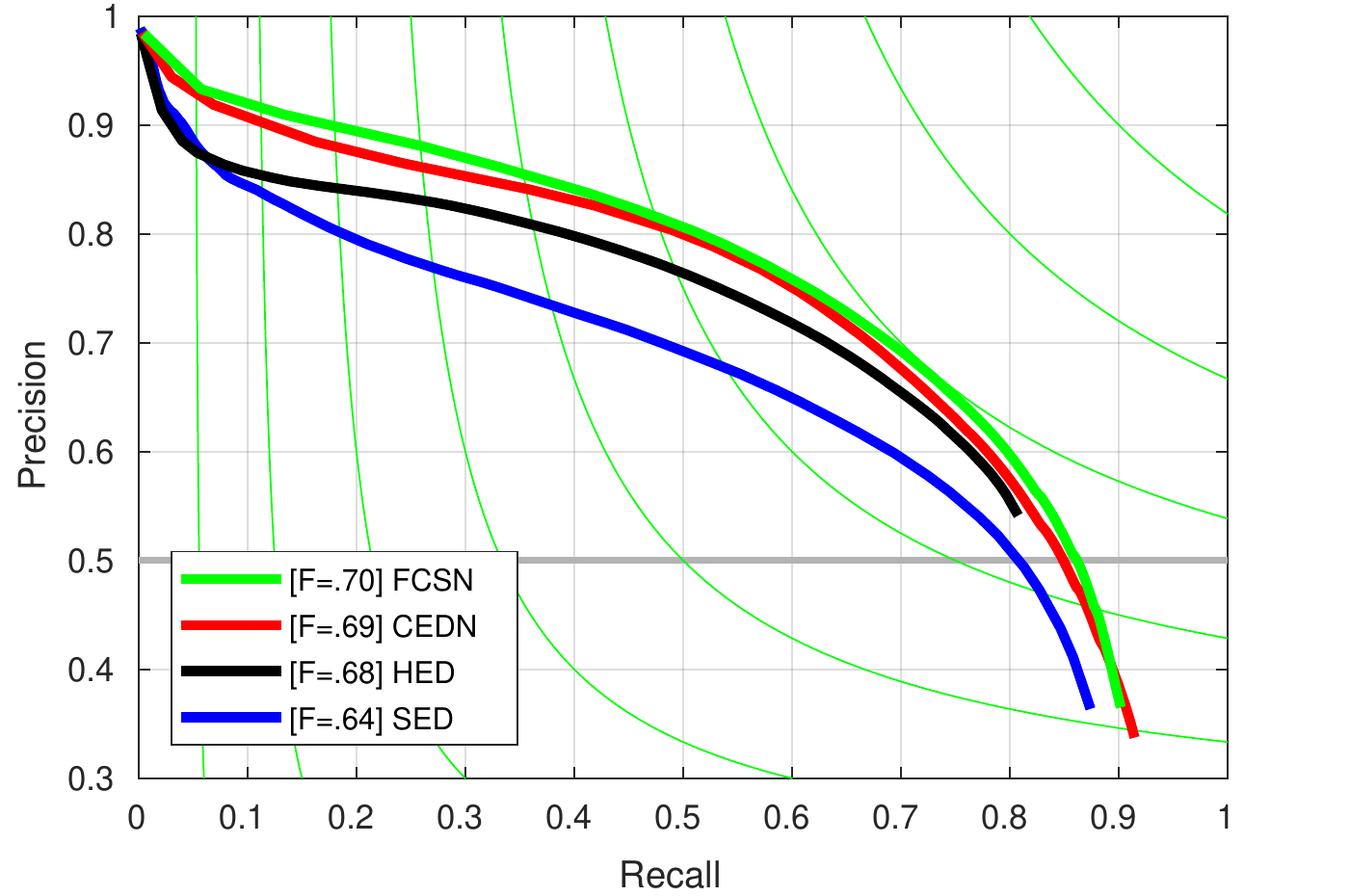}
                \caption{}
                \label{fig:fcsnvsbaselines}
        \end{subfigure}%
        \caption{(a) PR curve for object boundary detection on VSB100. (b) PR curve for object boundary detection on VSB100 with fine-tuning on both BSDS500 and VSB100 training sets.}\label{fig:animals}
\end{center}
\end{figure}

\noindent\textbf{Edgelet matching}. We apply edgelet-based matching to each edgelet pair, $e_t$ in frame $t$ and $e^{'}_{t+1}$ in frame $t+1$, 
that fall within a reasonable spatial neighborhood  (empirically set to 100 pixels around the edgelet as sufficient to 
accommodate for large motions). 
For each edgelet pair, $e_t$ in frame $t$ and $e^{'}_{t+1}$ in frame $t+1$, all
the similarities between all the pixel pairs on these edgelet
pairs are summed up and normalized to obtain the similarity
between the edgelet pair. The similarity between  
points $\langle\mathbf{x}_{t}^i,\mathbf{y}_{t+1}^j\rangle$ on $e_t$ and $e^{'}_{t+1}$ is  
expressed in terms of their respective excitation attention scores as $s_{ij}$.

For an edgelet $e_t$ in frame $t$, we keep the top-$10$ most similar edgelets in frame $t+1$ as its matching candidates. 
These candidate edgelet pairs are further filtered by their normals, with only edgelets with an angle not
more than $45$ degrees retained. The normals are computed
as the average direction from pixel coordinates on one side
of the edge to corresponding pixel coordinates on the other
side of the edge. This also helps to determine which superpixel pair falls on the same side of the edge in the two images. 
As shown in Fig.~\ref{fig:matching}(c), superpixels $s_1$ and $s_1^{'}$ fall on the left side of edges $e_t$ and $e_{t+1}^{'}$, respectively, thus superpixel pair $\{s_1, s_1^{'}\}$ fall on the same side of the edges.

After filtering by angle, a greedy matching algorithm
is performed to approximate bipartite matching of edgelets
in frame $t$ to edgelets in the frame $t+1$. This further reduces
the number of edgelet pairs retained.

For the final boundary flow placement, we observe that
some boundary points will be placing on the incorrect side
of the edgelet. We utilize normalized region similarity defined by color to assign the motion to superpixels pairs that are 
more similar to each other in color.
As shown in Fig.~\ref{fig:matching}(c), point $\mathbf{x}_t$ 
is on the right side of edge $e_t$ but the corresponding point $\mathbf{y}_{t+1}$ is on the left side of edge $e_{t+1}^{'}$. 
Our approach moves $\mathbf{x}_t$ to the other side of $e_t$, resulting in $\mathbf{x}_{t}^{new}$.
After moving the points, we obtain pixel-level matches which are the final boundary flow result.

\section{Training}
FCSN  is implemented using Caffe \cite{jia2014caffe}. 
The encoder weights are initialized with VGG-16 net and fixed during training. We update only the decoder parameters using the 
Adam method \cite{kingma2014adam} with learning rate $10^{-4}$. We train on VSB100, a state-of-the-art video object boundary dataset, which contains 40 training videos with annotations on every $20$-th frame, for a total of $240$ annotated frames. Because there are too few annotations, we augment the training with the PASCAL 
VOC 12 dataset, which contains 10582 still images (with refined object-level annotations as in \cite{yang2016object}). 
In each iteration, $8$ patches with size $224 \times 224$ are randomly sampled from an image pair of VSB100 (or two duplicated frames of PASCAL VOC) and passed to the model.

The loss function is specified as the weighted binary cross-entropy loss common in boundary detection ~\cite{xie2015holistically}
$
L(w){=}{-}\frac{1}{N}\sum_{i=1}^{N}[\lambda_1 y_n \log \hat{y}_{n}
+\lambda_2(1{-}y_n)\log (1{-}\hat{y}_{n})]
$
where N is the number of pixels in an iteration. Note that the loss is defined on a single side of the outputs, since only single frame annotations are available. 
The two decoder branches share the same architecture
and weights, and thus can be both updated simultaneously
with our one-side loss. The two branches still can output different predictions, since decoder predictions are modulated with different pooling indices recorded in the
corresponding encoder branches.
Due to the imbalances of boundary pixels and non-boundary pixels, we set $\lambda_1$ to $1$ and $\lambda_2$ to $0.1$, respectively.


\begin{table}
\centering
\begin{tabular}{|c|c|c|c|}
\hline

Method & 
ODS &
 OIS & 
 AP \\
\hline\hline
 CEDN \cite{yang2016object} & 0.563 &  0.614 &  0.547\\
\hline
 FCSN &   \bf 0.597 &  \bf 0.632 &  \bf 0.566\\
\hline
\end{tabular}
\caption{Results on VSB100.}
\label{table:cednvsfcsn}
\end{table}


\begin{table}
\centering
\begin{tabular}{|c|c|c|c|}
\hline
Method & ODS & OIS & AP \\
\hline\hline
SE \cite{Dollar2015PAMI} & $0.643$ & $0.680$ & $0.608$\\
\hline
HED \cite{xie2015holistically} & $0.677$ & $0.715$ & $0.618$\\
\hline
CEDN \cite{yang2016object} & $0.686$ & $0.718$ & $0.687$\\
\hline
FCSN & $\mathbf{0.698}$ & $\mathbf{0.729}$ & $\mathbf{0.705}$\\
\hline
\end{tabular}
\caption{Results on VSB100 with fine-tuning on both
BSDS500 and VSB100 training sets.}
\label{table:fcsnvsbaselines}

\end{table}

\begin{table}
\centering
\begin{tabular}{|c|c|c|c|c|}
\hline
Method & \begin{tabular}{@{}c@{}} FLANN \\ \cite{muja2009fast} \end{tabular} & \begin{tabular}{@{}c@{}} RANSAC \\ \cite{brown2003recognising} \end{tabular} & Greedy & \begin{tabular}{@{}c@{}}Our \\ Matching\end{tabular}\\
\hline
EPE & $23.158$ & $20.874$ & $25.476$ & $9.856$\\
\hline
\end{tabular}
\caption{Quantitative results of boundary flow on Sintel training dataset in EPE metric.}
\label{table:boundaryflow}
\end{table}


\section{Results}
This section presents our evaluation of boundary detection, BF estimation, and utility of BF for 
optical flow estimation.

\subsection{Boundary Detection}

After FCSN generates boundary predictions, we apply the standard non-maximum suppression (NMS). The resulting boundary detection is evaluated using precision-recall (PR) curves and F-measure.

\begin{figure}[htb]
\begin{center}
\includegraphics[width=1.0\linewidth]{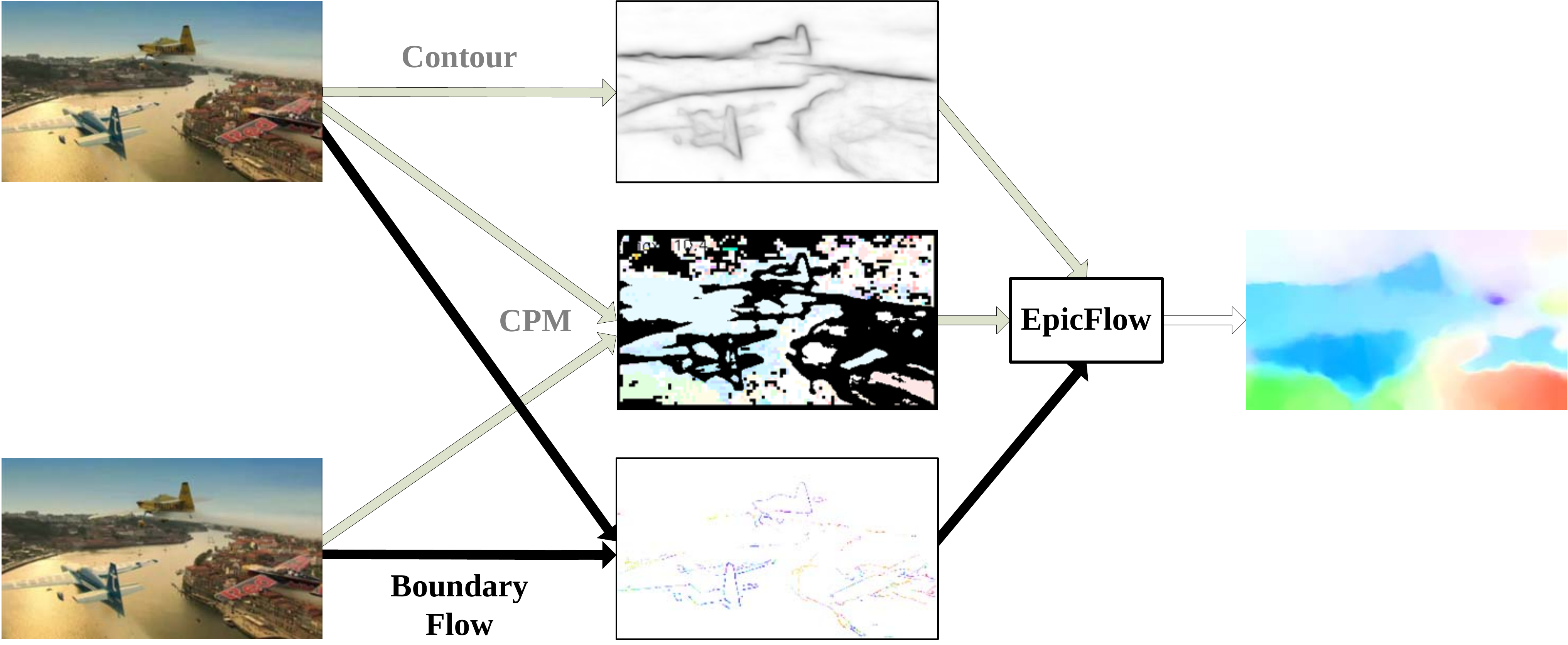}
\end{center}
\vskip -0.15in
   \caption{Overview of augmenting boundary flow into the framework of CPM-Flow. Given two images, we compute the standard input to CPM-Flow: matches using CPM matching \cite{huefficient16} and the edges of the first image using
SE \cite{Dollar2015PAMI}. Then we augment the matches with our predicted boundary flow (i.e., matches on the boundaries), as indicated by black arrows.}
\label{fig:aug}
\end{figure}

\begin{figure*}[t]
\begin{center}
\includegraphics[width=1.\linewidth]{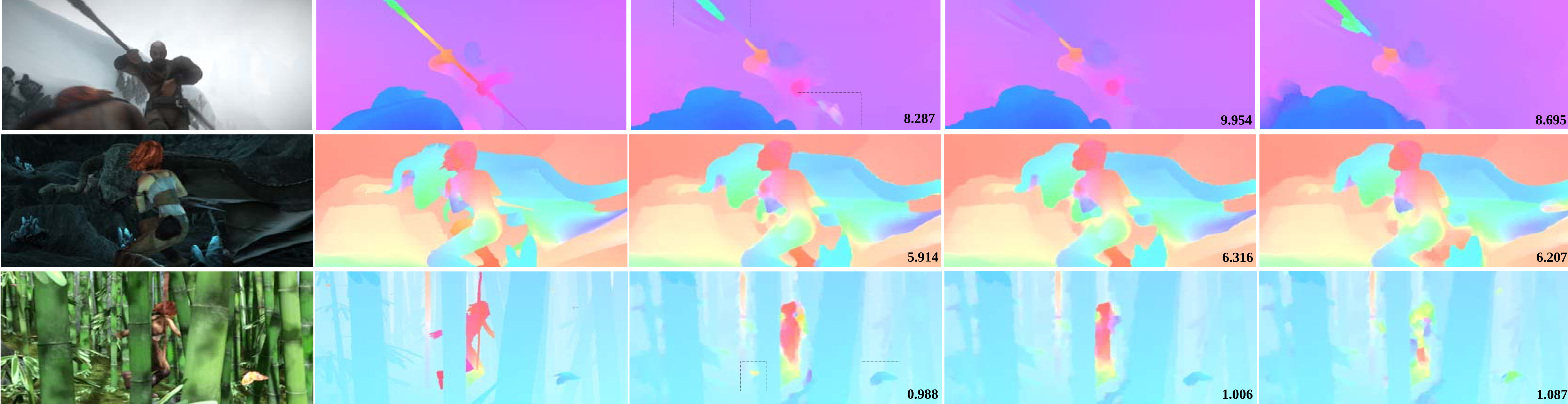}
\end{center}
\vskip -0.15in
   \caption{Example results on MPI-Sintel test dataset. 
The columns correspond to original images, ground
truth, CPM-AUG (i.e., our approach), CPM-Flow \cite{huefficient16} and EpicFlow\cite{revaud2015epicflow}. The rectangles highlight the improvements and the numbers indicate the EPEs. 
}
\label{fig:flow}
\end{figure*}

\noindent\textbf{VSB100.} For the benchmark VSB100 test dataset \cite{NB13}, we compare with the state-of-the-art approach CEDN \cite{yang2016object}. We train both FCSN and CEDN using the same training data with 30000 iterations. Note that CEDN is single-frame based. Nevertheless, both FCSN and CEDN use the same level of supervision, since only isolated single frame annotations apart from one another are available. Fig.~\ref{fig:cednvsfcsn} shows the PR-curves of object boundary detection. As can be seen, F-score of FCSN is $0.60$ while $0.56$ for CEDN. FCSN yields higher precision than CEDN, and qualitatively we observe that FCSN generates visually cleaner object boundaries. As shown in Fig.~\ref{fig:boundary}, CEDN misses some of the boundaries of background objects, but our FCSN is able to detect them. Due to limited training data,  both FCSN and CEDN obtain relatively low recall. Tab. \ref{table:cednvsfcsn} shows that FCSN outperforms CEDN in terms of the optimal dataset scale (ODS), optimal image scale (OIS), and average precision (AP).

\noindent\textbf{Finetuning on BSDS500 and VSB100.} We also evaluate another training setting when FCSN and CEDN are both fine-tuned on the BSDS500 training dataset \cite{arbelaez2011contour} and VSB100 training set for $100$ epochs with learning rate $10^{-5}$. BSDS has more edges annotated hence allows for higher recall. Such trained FCSN and 
CEDN are then compared with the state-of-the-art, including structured edge detection (SE) \cite{Dollar2015PAMI}, and  holistically-nested edge detection algorithm  (HED) \cite{xie2015holistically}. Both SE and HED are re-trained with the same training setting as ours. Fig.~\ref{fig:fcsnvsbaselines} and Tab.~\ref{table:fcsnvsbaselines} present the PR-curves and AP. As can be seen, FCSN outperforms CEDN, SE and HED in all metrics. We presume that further improvement may be obtained by training with annotated boundaries in both frames.

\subsection{Boundary Flow Estimation} 
Boundary flow accuracies are evaluated by average end-point error (EPE) between our boundary flow prediction 
and the ground truth boundary flow (as defined in Sec.~\ref{definition_of_bf}) on the Sintel training dataset.   

In order to identify a good competing approach, we have tested a number of the state-of-art matching algorithms on the Sintel training dataset, 
including coarse-to-fine PatchMatch (CPM) \cite{huefficient16}, Kd-tree PatchMatch \cite{he2012computing} and DeepMatching \cite{weinzaepfel2013deepflow}, 
but have found that these algorithms are not suitable for our comparison because they prefer to find point matches off  boundaries. 

Therefore, we compare our edgelet-based matching algorithm with the following baselines: (i) greedy nearest-neighbor point-to-point matching, 
(ii) RANSAC \cite{brown2003recognising}, (iii) FLANN, a matching method that uses SIFT feautres.
The quantitative results are summarized in Tab. \ref{table:boundaryflow}. Our edgelet-based matching outperforms all the baselines significantly.

\subsection{Dense Optical Flow Estimation}

We also test the utility of our approach for optical flow estimation on the Sintel testing dataset. After running our boundary flow estimation, the resulting boundary matches are used to 
augment the standard input to the state of the art CPM-Flow \cite{huefficient16}, as shown in Fig. \ref{fig:aug}. Such an approach is denoted as CPM-AUG, and compared with the other existing methods in Tab.~\ref{table:sinteltest}. 
As can be seen, CPM-AUG outperforms CPM-Flow and FlowFields.
Note the results we submitted on "Sintel clean"
under the name CPM-AUG was not actually results of
CPM-AUG, actually it was just our implementation of
CPMFlow\cite{huefficient16}, which is a bit lower than the public one on the Sintel dataset. 
However these are the best results we can obtain using the public implementation of the algorithm.
In principle, the augmented point matches should be able to help other optical flow algorithms as well as it is largely orthogonal to the information pursued by current optical flow algorithms.

Fig.~\ref{fig:flow} shows qualitative results of CPM-AUG on Sintel testing dataset with comparison to two state-of-the-art
methods: CPM-Flow and EpicFlow. As it can be seen, CPM-AUG performs especially well on the occluded
areas and benefits from the boundary flow to produce sharp motion boundaries on small objects like the leg and the claws as well as the elongated halberd.

\begin{table}
\centering
\begin{tabular}{|c|c|c|c|}
\hline
{Method} & \begin{tabular}{@{}c@{}}EPE \\ all\end{tabular} & \begin{tabular}{@{}c@{}}EPE \\ matched\end{tabular} & \begin{tabular}{@{}c@{}}EPE \\ unmatched\end{tabular}\\
\hline\hline
CPM-AUG & \textbf{5.645} & $2.737$ & \textbf{29.362}\\
\hline
FlowFields\cite{bailer2015flow} & $5.810$	& \textbf{2.621}	& $31.799$\\
\hline
Full Flow\cite{qifeng16cvpr} & $5.895$ & $2.838$ & $30.793$\\
\hline
CPM-Flow\cite{huefficient16} & $5.960$ & $2.990$& $30.177$\\
\hline
DiscreteFlow\cite{menze2015discrete} & $6.077$ & $2.937$ & $31.685$\\
\hline
EpicFlow\cite{revaud2015epicflow} & $6.285$ & $3.060$ & $32.564$\\
\hline
\end{tabular}
\caption{Quantitative results on Sintel final test set.}
\label{table:sinteltest}
\end{table}

\section{Conclusion}
We have formulated the problem of boundary flow estimation in videos. For this problem, 
we have specified a new end-to-end trainable FCSN which takes two images as input and 
produces boundary detections in each image. We have also used FCSN to generate excitation 
attention maps in the two images as informative features for boundary matching, thereby unifying 
detection and flow estimation. For matching points along boundaries, we have decomposed the predicted 
boundaries into edgelets and applied edgelet-based matching to pairs of edgelets from the two images. Our experiments on the 
benchmark VSB100 dataset for boundary detection demonstrate that FCSN is superior to the state- of-the-art, 
succeeding in detecting boundaries both of foreground and background objects. We have presented the 
first results of boundary flow on the benchmark Sintel training set, and compared with reasonable baselines. 
The utility of boundary flow is further demonstrated by integrating our approach 
with the CPM-Flow for dense optical flow estimation. This has resulted in an improved performance over the original CPM-Flow, especially on small details, sharp motion boundaries, and elongated thin objects in the optical flow. 

\noindent{\textbf{Acknowledgement}}. This work was supported in part by DARPA XAI Award N66001-17-2-4029.

{\small
\bibliographystyle{ieee}
\bibliography{egbib}
}

\end{document}